\PassOptionsToPackage{table,xcdraw}{xcolor}
\documentclass[table,xcdraw]{llncs} % 文档类会继承xcolor选项

\usepackage[T1]{fontenc}
\usepackage{graphicx}
\usepackage{booktabs}
\usepackage{caption}
\usepackage{makecell}
\usepackage[misc]{ifsym}
\usepackage{bbding}
\usepackage[most]{tcolorbox} % tcolorbox会继承已传递的xcolor选项
\usepackage{amsmath}
\usepackage{multirow}
\usepackage{diagbox}
\newcommand{\corr}{(\Letter)}

\usepackage{mwe}
\begin{document}

\title{LKD-KGC: Domain-Specific KG Construction via LLM-driven Knowledge Dependency Parsing
}
\newcommand{\name}{LKD-KGC }

\titlerunning{\name}

\author{
Jiaqi Sun\inst{1}\thanks{This work was completed during his internship at Alibaba Group.} \and
Shiyou Qian\inst{1} \corr \and
Zhangchi Han\inst{1} \and
Wei Li\inst{2} \and
Zelin Qian\inst{2} \and
Dingyu Yang\inst{3} \and
Jian Cao\inst{1} \and
Guangtao Xue\inst{1}
}

\authorrunning{J. Sun et al.}

\institute{
Shanghai Jiao Tong University, Shanghai, China 
\email{jotaro@sjtu.edu.cn, qshiyou@sjtu.edu.cn, hanzc\_2358@sjtu.edu.cn, caojian@sjtu.edu.cn, xue-gt@cs.sjtu.edu.cn}
\and
Alibaba Group, Hangzhou, Zhejiang, China 
\email{jianhao@taobao.com, zelin.qzl@alibaba-inc.com}
\and
Zhejiang University, Hangzhou, Zhejiang, China 
\email{yangdingyu@zju.edu.cn}
}

\maketitle              % typeset the header of the contribution

\begin{abstract}
Knowledge Graphs (KGs) structure real-world entities and their relationships into triples, enhancing machine reasoning for various tasks. While domain-specific KGs offer substantial benefits, their manual construction is often inefficient and requires specialized knowledge. 
Recent approaches for knowledge graph construction (KGC) based on large language models (LLMs), such as schema-guided KGC and reference knowledge integration, have proven efficient. However, these methods are constrained by their reliance on manually defined schema, single-document processing, and public-domain references, making them less effective for domain-specific corpora that exhibit complex knowledge dependencies and specificity, as well as limited reference knowledge.
To address these challenges, we propose LKD-KGC, a novel framework for unsupervised domain-specific KG construction. \name autonomously analyzes document repositories to infer knowledge dependencies, determines optimal processing sequences via LLM-driven prioritization, and autoregressively generates entity schema by integrating hierarchical inter-document contexts. This schema guides the unsupervised extraction of entities and relationships, eliminating reliance on predefined structures or external knowledge. Extensive experiments show that compared with state-of-the-art baselines, \name generally achieves improvements of 10\%–20\% in both precision and recall rate, demonstrating its potential in constructing high-quality domain-specific KGs.

\keywords{KG Construction  \and Large Language Model \and Domain-Specific KG \and Triple Extraction}

\end{abstract}

\section{Introduction}

%\begin{itemize}
%    \item task background
%    \item significance
%    \item features of domain-specific corpora
%    \item current work
%    \item limitations
%    \item method
%    \item contribution
%\end{itemize}

%说明知识图谱
%P1：知识图谱定义，重要性
%P1.5: KG提取的主要挑战
%P2: 为什么使用大模型提取KG
%P3: 当前工作：LLM提取KG的三条路线
%P4: 当前工作的不足：大模型仍然不能很好应对domain specific场景
%P5: 本文工作，如何应对这些挑战，框架
%P6: 实验效果
Knowledge Graph (KG) refers to a semantic network architecture that organizes information through graph-structured data made up of nodes and edges, formally characterizing attributes and relationships among real-world entities \cite{10.1145/3331166,KG-LLM-Survey}. It employs a triple-based representation of subject-predicate-object for systematic knowledge organization, which supports downstream applications that require complex inference.
For instance, domain-specific graph-based retrieval-augmented generation (Graph RAG) systems enable mapping natural language with relevant entities, providing structured knowledge support for question-answering systems  \cite{graphrag}. 

%P1.5: KG提取的主要挑战
Knowledge graph construction (KGC) involves extracting relational triples and their attributes from data sources, followed by organizing them into a coherent graph. 
However, manually constructing KGs proves to be inefficient due to the massive volume and high complexity of data  \cite{10.1145/3331166}. 
Domain-specific corpora further pose three unique challenges: (1) Complex knowledge dependency: Domain knowledge progresses hierarchically from foundational to advanced concepts, requiring cross-referential analysis; (2) Domain specificity: Containing dense usage of technical terminology, domain-specific abbreviations, and intricate implementation details; (3) Limited reference knowledge: Private technical documentation and operational case studies are usually unavailable through external knowledge sources.
These features place significant demands on the constructor's domain expertise  \cite{DBLP:journals/jnca/Abu-Salih21}.

%自动化知识图谱构建
%一笔带过传统方法和小模型，专注讨论大模型
%为什么使用LLM

Automated KGC methods are increasingly being developed to help knowledge extraction \cite{tang2024graphgptgraphinstructiontuning}. Early works based on traditional machine learning models  \cite{HMM,CRF} and earlier deep learning models  \cite{CNN,LSTM} either suffer from complex feature engineering or limited transferability across diverse tasks  \cite{mintz-etal-2009-distant}.
Recent methodologies propose leveraging large language models (LLMs) as automated KGC agents, which demonstrate strong capability in identification and structuring of complex relationships and entities from unstructured text  \cite{KG-LLM-Survey}.
LLMs overcome these limitations by leveraging self-supervised pre-training on massive text corpora to autonomously learn deep semantic patterns, while enabling zero/few-shot transfer across tasks through In-Context Learning (ICL) ability  \cite{10.5555/3495724.3495883}.

%统一介绍三类LLM-based的方法
Current research in LLM-based KGC primarily diverges into three directions: model fine-tuning, schema guidance and reference knowledge integration. 
\textbf{Model Fine-tuning:} Utilizing fine-tuned language models to generate structured triples directly from text, reducing manual efforts and improving contextual inference \cite{huguet-cabot-navigli-2021-rebel-relation,bosselut-etal-2019-comet}.
\textbf{Schema Guidance:} Employing predefined or dynamically generated schemas (metadata frameworks defining KG elements) to guide triple extraction and alignment \cite{EDC,RAP,CoT-Ontology,ye2023schemaadaptableknowledgegraphconstruction}.
\textbf{Reference Knowledge Integration:} Enhancing KGC by integrating external resources (search results, public KGs, domain documents) during  extraction \cite{AutoKG,SAC-KG,KBTE}.

%当前工作处理领域语料的不足
Despite advances in prior research, current LLM-based KGC methodologies still face challenges in processing domain-specific corpora, primarily in three aspects: the need for manually predefined schema, the dependence on external ground-truth knowledge, and the ignorance of knowledge dependency. 
For instance, the RAP framework requires a predefined schema to build schema-aware reference in prompt  \cite{RAP}, which is usually absent in real-world texts. 
AutoKG's real-time web retrieval mechanism becomes ineffective when handling internal system specification unavailable from public access \cite{AutoKG}.
Finally, while the EDC framework is effective at generating schema from individual documents, it is short at identifying cross-document entity relationships  \cite{EDC}.

%本文工作
To address these challenges, there is a growing need for frameworks that can handle the complexities of real-world data  \cite{ye2023schemaadaptableknowledgegraphconstruction}.
Inspired by the principle that knowledge advances from superficial to profound in real-world, we propose the LKD-KGC framework (\textbf{L}LM-dirven \textbf{K}nowledge \textbf{Dependency} Parsing for Domain-specific \textbf{K}nowledge \textbf{G}raph \textbf{C}onstruction), which fully utilizes hierarchical knowledge by considering knowledge dependencies.
\name basically consists of three components: Dependency Evaluation, Schema Definition, and Triple Extraction. 
\name first recursively accesses all directories and documents within the input repository, automatically evaluates document comprehension priorities through LLM to determine the optimal access order for the knowledge base, and generates context-aware document abstracts by incorporating this ordered contextual knowledge. Subsequently, it extracts all entity types and definitions from both original documents and their corresponding abstracts. Finally, based on the entity schema, \name identifies valid entities and their relationships within the original text, finally constructing a high-quality KG. 
%experiment results
The code implementation for LKD-KGC can be accessed via the following anonymous repository: \textit{https://anonymous.4open.science/r/KAD-KGC-DA17}.
Experimental results across three domain-specific datasets and two base LLMs demonstrate that compared with state-of-the-art baselines, our method generally achieves  improvements of 10\%–20\% in both precision and recall rate.

\section{Related Work}
%先总览三个类别
%每个类别：概述，举例，优缺点
As previously established, KGs present inherent construction challenges, while LLMs exhibit strong contextual semantic reasoning capabilities due to cross-domain generalizability from massive pre-training corpora \cite{KG-LLM-Survey}.
This complementary synergy drives three LLM-based KGC methodologies: model fine-tuning, schema guidance, and reference knowledge integration, which all exhibit limitations when applied to real-world domain-specific corpora.

\subsection{KGC based on Fine-tuned Language Models}
Previous KGC methods transition from multi-stage pipelines (entity discovery, coreference resolution, relation extraction) \cite{KG-LLM-Survey} to end-to-end generation via fine-tuned LLMs. COMET \cite{bosselut-etal-2019-comet} predicts tail entities given head entities and relations, while REBEL \cite{huguet-cabot-navigli-2021-rebel-relation} employs self-supervised learning for direct text-to-triple prediction, both leveraging generative language models like BART \cite{lewis-etal-2020-bart}. Despite outperforming fully supervised models like BiLSTM-CRF \cite{LSTM} in generalization, these approaches still face domain adaptation limitations.
Moreover, domain-specific corpora, such as operational cases in cloud systems, undergo rapid updates and lack sufficient annotation, further constraining the applicability of fine-tuned models. Consequently, emerging approaches increasingly leverage the ICL \cite{10.5555/3495724.3495883} capability of LLMs by incorporating diverse schemas and reference knowledge.

\subsection{KGC based on Schema Guidance}

Schemas (ontologies) formally define KG entities, relations, and attributes to ensure semantic coherence and quality in the constructed KG. Current approaches utilize LLMs' in-context learning through schema-guided prompts for triple extraction and alignment. 

For instance, CoT-Ontology \cite{CoT-Ontology} combines domain ontologies with Chain-of-Thought (CoT)  prompting for stepwise triples extraction.
RAP proposes a schema-aware retrieval-augmented approach that dynamically incorporates structured schema knowledge and semantically relevant instances as contextual prompts  \cite{RAP}. 
AdaKGC introduces schema-enriched prefix instructors and dynamic decoding methods to handle evolving schemas without requiring retraining model \cite{ye2023schemaadaptableknowledgegraphconstruction}. 
The Extract-Define-Canonicalize (EDC) framework breaks KGC into 3 stages: open information extraction, schema definition, and schema canonicalisation, which can be applied to scenarios whether a pre-defined target schema is available or not  \cite{EDC}.

These methods have limitations when processing domain-specific corpora, primarily manifested in two aspects: dependence on manually defined schemas and neglect of knowledge dependency. Specifically, both CoT-ontology and RAP depend on predefined schemas to refine prompts. 
While AdaKGC enables dynamic schema adjustment during KG construction or maintenance, it cannot create schemas from scratch. 
Although EDC can dynamically construct schemas based on input texts, its schema definition phase depends solely on relationships identified within individual documents, thereby missing inter-document conceptual dependencies.

\subsection{KGC based on Reference Knowledge Integration}
This methodology enhances KGC through multi-source knowledge integration (search engines, open-source KGs, domain corpora), equipping LLMs with contextualized domain awareness to improve triple extraction accuracy and fidelity.

For example, the SAC-KG framework employs a generator that
retrieves relevant context and triples for specified entities using domain corpora and DBpedia  \cite{SAC-KG}. 
TKGCon develops an automatic framework to produce fine-grained theme KGs using theme schemas derived from Wikipedia categories  \cite{wiki_as_resource}, thus reducing human effort and keeping up with changing knowledge  \cite{ding2024automatedconstructionthemespecificknowledge}. 
AutoKG employs communicative intelligent agents assigned with different
roles to work together on KGC, while incorporating real-time Internet retrieval systems  \cite{AutoKG}.
Papaluca et al. propose the KBTE pipeline that dynamically gathers contextual information from a Knowledge Base (KB) derived from input corpora, both in the form of context triplets and of <sentence, triplets> pairs as few-shot examples  \cite{KBTE}. 

These methods face limitations because of reliance on external knowledge. SAC-KG requires seed entities and reference triples from open-source KG, TKGCon employs Wikipedia taxonomies, AutoKG relies on web retrieval, and KBTE constructs its knowledge base from human-annotated ground truth triples.
However, for private corpora such as internal enterprise system specifications or operational cases in cloud systems, significant challenges arise due to domain-specific terminologies, concepts, and abbreviations known only to internal personnel. These resources remain inaccessible through public networks and lack annotated training data with ground truth triples.

\section{The \name Framework}
%P1: 设计思想
%P2: 模块，工作流程
%P3.1~3.3: 具体模块介绍

\begin{table}[tbp]
\renewcommand{\arraystretch}{1.2} 
\captionsetup{skip=5pt}
\setlength{\tabcolsep}{7pt}  
\caption{Comparison between \name and recent LLM-based KGC methods} \label{comparision}
\begin{tabular}{|c|c|c|c|c|}
\hline
Method & \begin{tabular}[c]{@{}c@{}}Independent \\ of Predefined \\ Schema\end{tabular} & \begin{tabular}[c]{@{}c@{}}Independent of \\ Ground Truth \\ Knowledge\end{tabular} & \begin{tabular}[c]{@{}c@{}}Global \\ Knowledge\\ Context\end{tabular} & \begin{tabular}[c]{@{}c@{}}Knowledge \\ Dependency\\ Awareness\end{tabular} \\ \hline
EDC \cite{EDC} & \CheckmarkBold & \CheckmarkBold & \XSolidBrush & \XSolidBrush \\ \hline
AutoKG \cite{AutoKG} & \CheckmarkBold & \XSolidBrush & \XSolidBrush & \XSolidBrush \\ \hline
KBTE \cite{KBTE} & \CheckmarkBold & \XSolidBrush & \XSolidBrush & \XSolidBrush \\ \hline
SAC-KG \cite{SAC-KG} & \CheckmarkBold & \XSolidBrush & \CheckmarkBold & \XSolidBrush \\ \hline
RAP \cite{RAP} & \XSolidBrush & \XSolidBrush & \CheckmarkBold & \XSolidBrush \\ \hline
CoT-Ontology \cite{CoT-Ontology} & \XSolidBrush & \CheckmarkBold & \CheckmarkBold & \XSolidBrush \\ \hline
\name(Ours) & \CheckmarkBold & \CheckmarkBold & \CheckmarkBold & \CheckmarkBold \\ \hline
\end{tabular}
\end{table}

\subsection{Overview}
\subsubsection{Design principles}
%插入和当前工作的最大不同（dependency）,说明表1
\name uniquely prioritizes global knowledge dependencies through hierarchical knowledge access, inspired by the nature that human acquire knowledge from foundational to advanced concepts. This methodology processes corpora while leveraging contextual knowledge from preceding corpora segments.
Table \ref{comparision} demonstrates the comparison between \name and recent LLM-based KGC methods, where only \name does not rely on predefined schemas or external reference knowledge, while addressing Global Knowledge Context with knowledge dependency awareness.

\begin{figure}[tbp]
\centering
\includegraphics[width=1.0\textwidth]{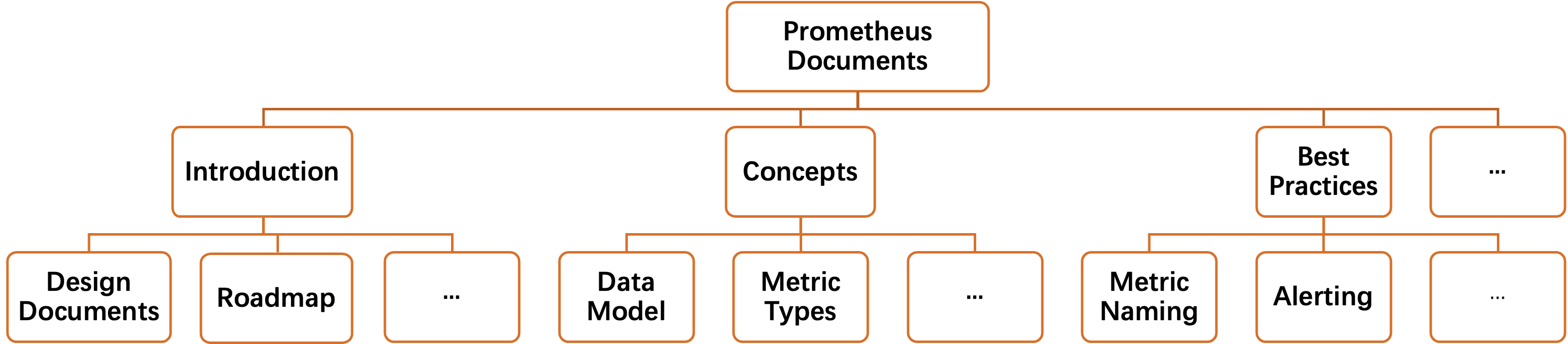}
\caption{The directory structure of Prometheus documents} \label{prom_docs}
\end{figure}

\subsubsection{Observation}
The design of \name is derived from two critical observations pertaining to domain-specific corpora.
\begin{enumerate}
    %观察1:领域知识的访问顺序
    \item Domain corpora exhibit latent ideal accessing orders. 
    Taking the popular open-source monitoring system Prometheus \cite{prometheus} as an example, as demonstrated in Figure \ref{prom_docs}, Prometheus' documentation \cite{prom_docs}, forms a directory tree with documents as leaves.
    When constructing knowledge graphs from this repository, initially processing the text within the \textit{Introduction} directory helps understanding of the system's overall architecture. This should be followed by processing the content in \textit{Concepts} to comprehend system components and terminology, before finally addressing various downstream application cases in \textit{Best Practices}. 
    
    %观察2: 摘要有助于总结实体类型和关系
    \item Text summarization aids in condensing entity types. By identifying and condensing key entities, it emphasizes semantically significant elements and relations while eliminating redundancy.
    For instance, when utilizing Llama-3.1-70B-Instruct \cite{llama} to summarize Prometheus' \textit{Metric Types} document, it generates a summary as shown in the textbox below. By providing both the summary and the original text, LLMs can accurately summarize the four primary entity types: Counter, Gauge, Histogram, and Summary, while avoiding redundant types.
\end{enumerate}

\begin{tcolorbox}[
  title=Example of Document Summarization for \textit{Metric Types}, 
  colback=blue!5!white, % 背景色
  colframe=blue!10!white,, % 边框颜色
  coltitle=black, % 标题颜色
  fonttitle=\bfseries
]
 Prometheus client libraries offer four core metric types: Counter, Gauge, Histogram, and Summary. These types are differentiated in client libraries and the wire protocol, but the Prometheus server currently flattens all data into untyped time series.
 %The document provides detailed definitions and use cases for each metric type, including examples and links to client library usage documentation. It also highlights the structure and functionality of Histograms and Summaries, noting experimental support for native histograms in Prometheus v2.40. The content emphasizes the appropriate use of each metric type and their roles in monitoring and data collection.

\end{tcolorbox}

%观察2:文本摘要有助于总结实体类型。

%观察发现，文档的层次结构
%进一步发现，摘要有助于总结实体类型
%以IMS为例

\subsubsection{Workflow}
\begin{figure}[t]
\centering
\includegraphics[width=1\textwidth]{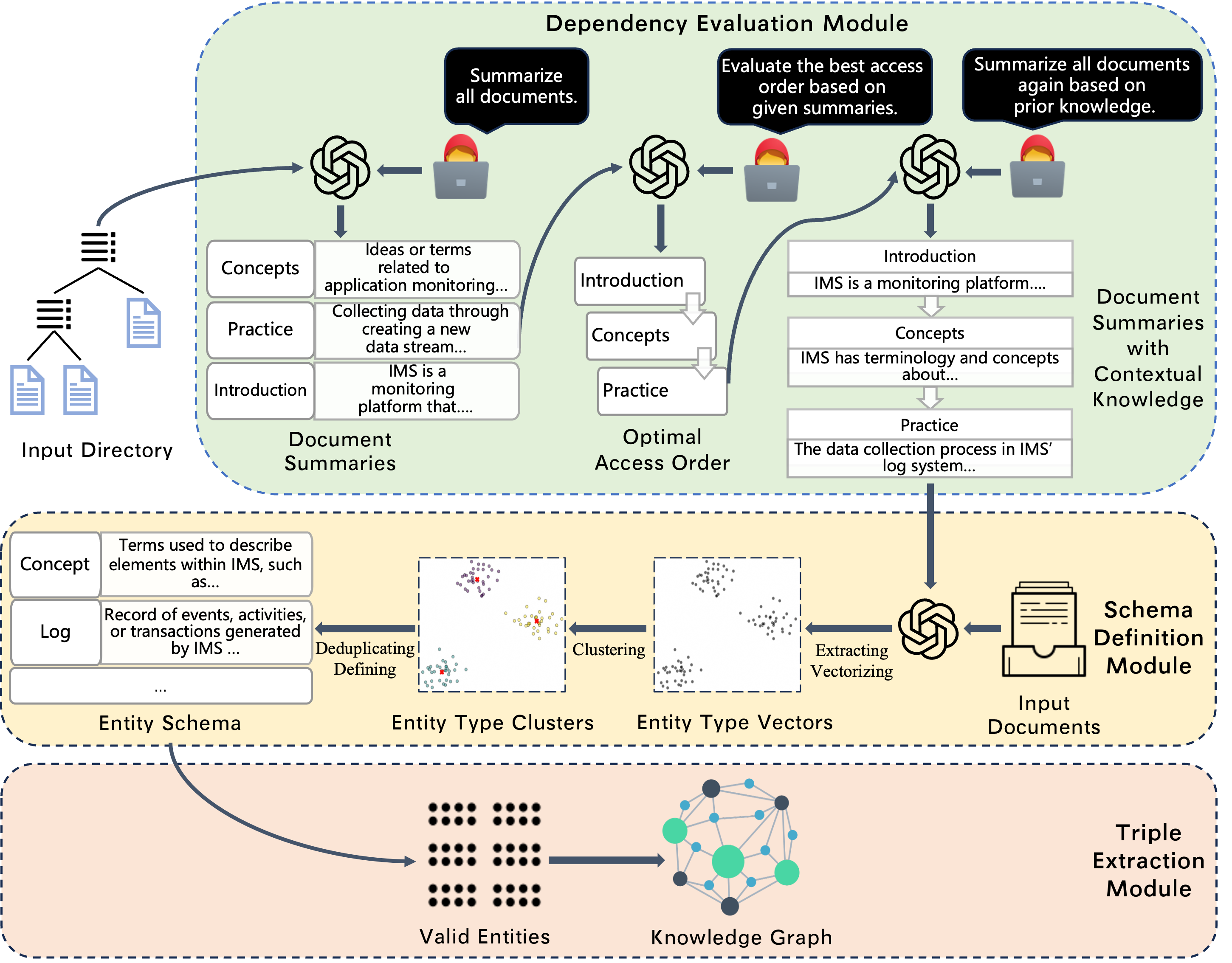}
\caption{The framework of LKD-KGC. We employ the system specifications of an internal application monitoring system of a large Internet company as an illustrative example.
To preserve author anonymity, we adopt the codename "IMS"(Internal Monitoring System) for referencing this corpus.
There is almost no publicly available information regarding this system on the Internet.} \label{framework}
\end{figure}

As depicted in Figure \ref{framework}, the \name framework comprises three modules for Dependency Evaluation, Schema Definition, and Triple Extraction.
The modules sequentially implement a three-stage workflow: evaluating optimal knowledge access order → constructing entity schema → extracting triples under the guidance of established entity schema.

\begin{enumerate}
    \item The Dependency Evaluation Module begins with a bottom-up traversal of the tree to generate summaries for documents or directories corresponding to each node. It then performs a top-down recursive ordering from the root node, assessing the optimal access order. Finally, it generates summaries for each document, incorporating the summaries of previously ordered documents as contextual knowledge.
    \item For each document, the Schema Definition Module extracts entity types and their definitions from the original text and its summary. These entity types are embedded as semantic vectors and clustered. Similar entity types in each cluster are unified to retrieve relevant summaries and annotated with definitions.
    \item The Triple Extraction Module extracts valid entities from the original text based on specified entity types and definitions, and subsequently identifies relationships among these valid entities to construct the final KG.
\end{enumerate}

%三个模块分开
%只展示模块输出结果，不用详细写每个操作
%展示具体示例

\subsection{System Components}
%dependency evaluation module, schema definition module and triple extraction module
%首先整体说明模块的作用
\subsubsection{Dependency Evaluation Module}
%将目录视为树结构，首先遍历树，为目录的叶子结点的文档生成摘要
%有目录结构的情况下，不但为文档生成摘要，目录节点也有摘要，然后由根节点向下递归排序
This module is designed to determine the optimal order for knowledge access and guide the generation of context-aware document summaries. When processing long unstructured text without a directory structure, we can split the text into chunks, treating each chunk as an individual document to establish a single-layer flattened structure. Therefore, we assume a more complex scenario where the input document repository contains multi-layered hierarchical structures.

%自下而上摘要
When processing a document root directory, the Dependency Evaluation Module first traverses the entire directory tree and prompts the model to generate summaries for each document. This initial step aims to understand document content as the basis for ordering, requiring no specific access order and generating summaries without cross-document references. After obtaining summaries of each individual document, the module hierarchically generates directory summaries in a bottom-up sequence, finally establishing summaries containing key entities and relationships for each node in the directory tree.

%自上而下遍历
Subsequently, the module performs a top-down traversal of the directory tree. At each hierarchical level, it constructs prompts using existing summaries to assess optimal access order. It recursively applies this evaluation process when encountering subdirectories. For leaf-node documents, it generates context-enhanced summaries by incorporating historical knowledge from previously accessed documents into prompts. This autoregressive approach ensures that each document summary is generated with reference to prior knowledge, progressively building a comprehensive knowledge repository through ordered knowledge integration.

%长上下文的情况
In case of processing large volumes of data, including all prior document summaries in prompts becomes infeasible due to model context length constraints. To address this issue, we implement a vector retrieval functionality: each document summary is stored in a vector database with its embedding as the key, and the original summary text as the value. As processing subsequent documents, the content of new documents is vectorized as query vectors to retrieve the top-$k$ related summaries ($k=10$ by default) exhibiting the smallest cosine distance (i.e., highest semantic similarity) from the vector database. This RAG-like approach enables adaptation of referenced knowledge context length to LLM's context length.
%流程图,具体实例

\subsubsection{Schema Definition Module}
This module extracts entity types from individual documents based on text summaries containing contextual information. It then performs entity type clustering and deduplication, ultimately generating definition for each entity type, as the final entity schema. Specifically, the module first constructs a prompt for each document (as illustrated below, taking \textit{Prometheus Metric Types} as an example), instructing the LLM to summarize all meaningful entity types.

\begin{tcolorbox}[
  title=\small{Example of Entity Type Extraction Prompt},
  colback=blue!5!white, % 背景色
  colframe=blue!10!white, % 边框颜色
  coltitle=black, % 标题颜色
  fonttitle=\bfseries,
  breakable=true
]
\textbf{Task}: Some text and its summary will be provided. You need to refer to the summary and the content of the text to classify meaningful entities within the text, and return a list containing all the entity types.\\
\textbf{Text}: A counter is a cumulative metric that... \\
\textbf{Summary}:  Prometheus client libraries offer four core metric types: Counter, Gauge, Histogram, and Summary...\\
------------------------------------------------------------------------------------------\\
\textbf{LLM}: [Counter, Gauge, Histogram, Summary]
\end{tcolorbox}

However, the initially extracted entity types usually exhibit an excessive quantity with significant redundancy. To resolve coreference and deduplicate entity types, this module first  filters out entity types that appear only once. Then, it vectorizes all entity types to embeddings and aggregates semantically similar ones through K-means clustering algorithm \cite{kmeans}, as formalized below:
\begin{equation}
\arg\min_{S} \sum_{j=1}^{k} \sum_{\mathbf{x}_i \in C_j} \|\mathbf{x}_i - \boldsymbol{\mu}_j\|^2,
\end{equation}
where $\mathbf{x}_i $ is the embedding of the $i$-$th$ entity type, and $S = \{ C_1, C_2, \dots, C_k \}$ is the collection of all entity type clusters. 
Here cluster number $k$ is a hyperparameter, and \name automatically determines the optimal $k$ by silhouette coefficient $s$ \cite{silhouettes}, which is a quantitative measure that evaluates clustering quality. A higher $s$ indicates better clustering quality, with stronger intra-cluster cohesion and clearer separation between clusters:
\begin{equation}
\arg\max_{k} \quad  \frac{1}{N} \sum_{i=1}^{N} \frac{b(i) - a(i)}{\max({a(i), b(i)})},
\end{equation}
where $a(i)$ = mean intra-cluster distance of entity type vector $i$ (compactness); $b(i)$= mean nearest-cluster distance of entity type vector $i$  (separation).

Only entity types remain insufficient to constitute a complete entity schema. Therefore, in the final stage, the Schema Definition Module presents each entity type cluster to LLM, directing it to deduplicate semantically equivalent types and formulate comprehensive definitions for all entity types. During this process, vector retrieval is performed with the summary knowledge base to recall summaries relevant to each cluster. These retrieved summaries serve as contextual knowledge to facilitate the model's entity definition as illustrated below. Finally, the Schema Definition Module produces a complete entity schema containing refined and meaningful entity types with their definitions.

\begin{tcolorbox}[
  title=\small{Example of Entity Type Definition Prompt},
  colback=blue!5!white, % 背景色
  colframe=blue!10!white, % 边框颜色
  coltitle=black, % 标题颜色
  fonttitle=\bfseries,
  breakable=true
]
\textbf{Task}: Here are some entity types.  First, merge types that have the same meaning. Then, briefly define each merged type in one sentence. These entity types come from the same knowledge base; you can refer to the reference knowledge below.\\
\textbf{Input entity types}: [Configuration, Configures, ...] \\
\textbf{Reference Knowledge}: "guides/multi-target-exporter": The document provides a YAML configuration example for setting up Prometheus to scrape metrics from a Blackbox Exporter...\\
------------------------------------------------------------------------------------------\\
\textbf{LLM}: \{"Configuration": Prometheus is configured by YAML files that define scrape targets, intervals, storage settings, and alerting rules, ...\}
\end{tcolorbox}

\subsubsection{Triple Extraction Module}
This module prompts LLM to extract valid relation triples from the original text under the guidance of the entity schema. Given the large number of entity types (e.g., 66 entity types extracted from 62 Prometheus documentation articles using Llama-3.1-70B-Instruct \cite{llama}), the potential relationship types between entity types grow exponentially, making the definition of all relation types impractical.
Therefore, the Triple Extraction Module does not define relation types. Instead, it first extracts valid entities and then extracts potential relations based on given entities as illustrated below, which finally form a comprehensive KG.

\begin{tcolorbox}[
  title=\small{Example of Triple Extraction},
  colback=blue!5!white, % 背景色
  colframe=blue!10!white, % 边框颜色
  coltitle=black, % 标题颜色
  fonttitle=\bfseries,
  breakable=true
]
\textbf{Task}: Here is a document. Please extract meaningful entities and their types which belongs to the given entity schema from the document.\\
\textbf{Document content}: This guide will introduce you to the multi-target exporter pattern...\\
\textbf{Entity Schema}: \{"Configuration": Prometheus is configured by YAML files that define scrape targets, intervals, and alerting rules, ...\}\\
------------------------------------------------------------------------------------------\\
\textbf{LLM}: \{Configuration:[Blackbox.yml, Prometheus.yml], ...\}\\
------------------------------------------------------------------------------------------\\
\textbf{Task}: Below is a document and some entities extracted from it. What entities do you believe have clear and meaningful relationships? Give all relation triples in the format of [(Subject Entity, Predicate, Object Entity)].\\
\textbf{Document content}: This guide will introduce you to the multi-target exporter pattern...\\
\textbf{Entities} \{Configuration:[Blackbox.yml, Prometheus.yml], ...\}\\
\textbf{Entity Schema}: \{"Configuration": Prometheus is configured by YAML files that define scrape targets, intervals, and alerting rules, ...\}\\
------------------------------------------------------------------------------------------\\
\textbf{LLM}: [(Prometheus.yml, Defines, Scrape\_configs),...]
\end{tcolorbox}

\section{Experiments}
In this section, we compare \name with three LLM-based methods that unsupervisedly extract knowledge graphs from open texts. Experiments across three datasets and two base LLMs demonstrate that \name outperforms state-of-the-art methods in both precision and recall metrics.
\subsection{Experimental Setup}
\subsubsection{Datasets}
We select three domain-specific corpora, including two public corpus and one private corpus, as follows:
\begin{enumerate}
    \item \textbf{Prometheus documentation \cite{prom_docs}}: This corpus comprises 62 documents detailing the architecture, concepts, and tutorials of the Prometheus monitoring system \cite{prometheus}, with extensive publicly available reference materials.
    
    \item \textbf{Re-DocRED} \cite{re-docred}: A refined version of DocRED \cite{yao-etal-2019-docred} (the most popular benchmark for document-level relation extraction), enhanced through re-annotation of omitted relation triples. We extracted a Windows-centric subset containing 15 documents, like \textit{Windows Operating Systems} and \textit{Windows Softwares}.

    \item \textbf{IMS documentation}:  Containing 46 internal technical documents from a large Internet company, covering product specifications, terminologies, and guidelines for a monitoring system. There are almost no reference knowledge about this system on the Internet.
\end{enumerate}

\noindent\textbf{Baselines}\quad
%\subsubsection{Baselines}
We selected three LLM-based methods that extract knowledge graph without supervision as baselines: EDC \cite{EDC}, AutoKG \cite{AutoKG}, and KBTE \cite{KBTE}. All of these approaches do not rely on manually predefined schemas.
Among them, KBTE requires constructing a knowledge base using ground truth tuples from the training set, which are unavailable in the open texts. Therefore, during reproduction, we build the knowledge base following KBTE's methodology based on tuples extracted from preceding documents.

\noindent\textbf{Experiment Configurations}\quad
%\subsubsection{Experiment Configurations}
We employ two base LLMs: Llama-3.1-70B-Instruct \cite{llama} and DeepSeek-R1-Distill-Qwen-14B \cite{deepseek-r1}. Following previous work, the temperature parameter for LLM inference is set to 0.1 to control output stability \cite{SAC-KG}. The context length of vector retrieval is set to 10 summaries. Finally, the text embedding model used in all experiments is paraphrase-multilingual-MiniLM-L12-v2 \cite{reimers-2019-sentence-bert}.

\noindent\textbf{Evaluation Metrics}\quad
%\subsubsection{Evaluation Metrics}
To evaluate the quality of KGs, we assess the precision and recall of extracted triples. 
Specifically, the Prometheus and IMS datasets lack manually annotated ground-truth triples, making direct validation of triple correctness infeasible. Given the substantial volume of triples, human evaluation also proves impractical. Following Zheng et al.'s finding that LLMs achieve human-comparable performance in fact assessment when combined with source text \cite{llm-as-judge,SAC-KG}, we adopt the LLM-as-judge approach from prior work, inputting extracted triples alongside their corresponding text into Qwen2-VL-72B \cite{qwen2} to evaluate precision. Calculating recall rate is infeasible due the absence of ground-truth tuples. Following Chen et al., we instead compute the number of true triples per document as recall metric, which aligns with real-world scenarios \cite{SAC-KG}.

The Re-DocRED dataset provides ground-truth triples. However, due to the absence of training process, the outputs of LLMs fail to strictly align with the ground truth, exhibiting numerous entities and relations that share same semantics but differ in expressions (e.g., "Windows" vs. "Microsoft Windows"). We therefore employ the LLM-as-judge methodology again, utilizing LLMs to assess semantic equivalence between extracted triples and ground truth triples, subsequently computing precision, recall, and F1 score.

\begin{table}[t]
% \footnotesize
\renewcommand{\arraystretch}{1.2} 
\captionsetup{skip=5pt}
\setlength{\tabcolsep}{5pt} 
\caption{Evaluation results on two public datasets: Prometheus Documentations(\uppercase\expandafter{\romannumeral1}) and Re-DocRED(\uppercase\expandafter{\romannumeral2})} \label{public_res}
\begin{tabular}{|cc|ccc|ccc|}
\hline
\rowcolor[HTML]{656565} 
\multicolumn{2}{|c|}{\cellcolor[HTML]{656565}{\color[HTML]{FFFFFF} \diagbox[linecolor=darkgray]{Dataset}{Model} }} 

&
  \multicolumn{3}{c|}{\cellcolor[HTML]{656565}{\color[HTML]{FFFFFF} Meta-Llama-3.1-70B-Instruct}} &
  \multicolumn{3}{c|}{\cellcolor[HTML]{656565}{\color[HTML]{FFFFFF} DeepSeek-R1-Distill-Qwen-14B}} \\ \hline
\rowcolor[HTML]{EFEFEF} 
\multicolumn{1}{|c|}{\cellcolor[HTML]{EFEFEF}} &
  %\diagbox{Method}{Metric} &
  Method\textbf{\textbackslash}Metric &
  \multicolumn{1}{c|}{\cellcolor[HTML]{EFEFEF}Precision} &
  \multicolumn{1}{c|}{\cellcolor[HTML]{EFEFEF}\begin{tabular}[c]{@{}c@{}}Recall\\ Number\end{tabular}} &
  \begin{tabular}[c]{@{}c@{}}Average\\ Recall \\ Number\end{tabular} &
  \multicolumn{1}{c|}{\cellcolor[HTML]{EFEFEF}{\color[HTML]{000000} Precision}} &
  \multicolumn{1}{c|}{\cellcolor[HTML]{EFEFEF}\begin{tabular}[c]{@{}c@{}}Recall\\ Number\end{tabular}} &
  \begin{tabular}[c]{@{}c@{}}Average\\ Recall \\ Number\end{tabular} \\ \cline{2-8} 
\multicolumn{1}{|c|}{\cellcolor[HTML]{EFEFEF}} &
  EDC &
  \multicolumn{1}{c|}{65.4\%} &
  \multicolumn{1}{c|}{3784} &
  61 &
  \multicolumn{1}{c|}{69.3\%} &
  \multicolumn{1}{c|}{3870} &
  62.4 \\ \cline{2-8} 
\multicolumn{1}{|c|}{\cellcolor[HTML]{EFEFEF}} &
  AutoKG &
  \multicolumn{1}{c|}{65.7\%} &
  \multicolumn{1}{c|}{2678} &
  43.2 &
  \multicolumn{1}{c|}{66.3\%} &
  \multicolumn{1}{c|}{2138} &
  34.5 \\ \cline{2-8} 
\multicolumn{1}{|c|}{\cellcolor[HTML]{EFEFEF}} &
  KBTE &
  \multicolumn{1}{c|}{75.6\%} &
  \multicolumn{1}{c|}{2216} &
  35.7 &
  \multicolumn{1}{c|}{63.0\%} &
  \multicolumn{1}{c|}{1503} &
  24.2 \\ \cline{2-8} 
\multicolumn{1}{|c|}{\multirow{-7}{*}{\cellcolor[HTML]{EFEFEF}\begin{tabular}[c]{@{}c@{}} \uppercase\expandafter{\romannumeral1}\end{tabular}}} &
  LKD-KGC &
  \multicolumn{1}{c|}{\textbf{83.4\%}} &
  \multicolumn{1}{c|}{\textbf{4561}} &
  \textbf{73.6} &
  \multicolumn{1}{c|}{\textbf{84.7\%}} &
  \multicolumn{1}{c|}{\textbf{4047}} &
  \textbf{65.3} \\ \hline
\rowcolor[HTML]{EFEFEF} 
\multicolumn{1}{|c|}{\cellcolor[HTML]{EFEFEF}} &
  %{\color[HTML]{000000} \diagbox{Method}{Metric}} &
  Method\textbf{\textbackslash}Metric &
  \multicolumn{1}{c|}{\cellcolor[HTML]{EFEFEF}{\color[HTML]{000000} Precision}} &
  \multicolumn{1}{c|}{\cellcolor[HTML]{EFEFEF}{\color[HTML]{000000} Recall}} &
  {\color[HTML]{000000} \begin{tabular}[c]{@{}c@{}}F1 \\ Socre\end{tabular}} &
  \multicolumn{1}{c|}{\cellcolor[HTML]{EFEFEF}{\color[HTML]{000000} Precision}} &
  \multicolumn{1}{c|}{\cellcolor[HTML]{EFEFEF}{\color[HTML]{000000} Recall}} &
  {\color[HTML]{000000} \begin{tabular}[c]{@{}c@{}}F1 \\ Score\end{tabular}} \\ \cline{2-8} 
\multicolumn{1}{|c|}{\cellcolor[HTML]{EFEFEF}} &
  EDC &
  \multicolumn{1}{c|}{17.7\%} &
  \multicolumn{1}{c|}{14.6\%} &
  16 &
  \multicolumn{1}{c|}{10.3\%} &
  \multicolumn{1}{c|}{16.5\%} &
  12.6 \\ \cline{2-8} 
\multicolumn{1}{|c|}{\cellcolor[HTML]{EFEFEF}} &
  AutoKG &
  \multicolumn{1}{c|}{22.8\%} &
  \multicolumn{1}{c|}{20.9\%} &
  20.9 &
  \multicolumn{1}{c|}{23.9\%} &
  \multicolumn{1}{c|}{16.9\%} &
  19.8 \\ \cline{2-8} 
\multicolumn{1}{|c|}{\cellcolor[HTML]{EFEFEF}} &
  KBTE &
  \multicolumn{1}{c|}{32.3\%} &
  \multicolumn{1}{c|}{18.8\%} &
  23.8 &
  \multicolumn{1}{c|}{25.9\%} &
  \multicolumn{1}{c|}{22.0\%} &
  23.8 \\ \cline{2-8} 
\multicolumn{1}{|c|}{\multirow{-6}{*}{\cellcolor[HTML]{EFEFEF}\uppercase\expandafter{\romannumeral2}}} &
  LKD-KGC &
  \multicolumn{1}{c|}{\textbf{36.3\%}} &
  \multicolumn{1}{c|}{\textbf{25.1\%}} &
  \textbf{29.7} &
  \multicolumn{1}{c|}{\textbf{32.2\%}} &
  \multicolumn{1}{c|}{\textbf{27.1\%}} &
  \textbf{29.5} \\ \hline
\end{tabular}
\end{table}

\subsection{Evaluation Results}
\subsubsection{Public Corpora}
We first conducted evaluations on two public datasets. Experimental results shown in Table \ref{public_res} demonstrate that \name achieves significantly higher precision and greater numbers of recalled triples than all baselines on both Meta-Llama-3.1-70B-Instruct and DeepSeek-R1-Distill-Qwen-14B. \\

\noindent\textbf{Prometheus Documentations(\uppercase\expandafter{\romannumeral1})}\quad
In terms of precision, the three baselines demonstrated comparable performance. However, \name significantly reduces the false positive rate and achieves higher accuracy by first extracting valid entities according to entity schema, then discovering relationships based on these valid entities.

As for recall number, due to the highly specialized nature of Prometheus documentation that contains a vast number of technical terms, \name extracted 6,580 valid entities, averaging 110.5 per document, thereby discovering more valid relations and achieving an advantage in recall number. Among the three baselines, EDC achieved significantly higher recall numbers. This difference primarily stems from the baselines' operational scopes: EDC performs open information extraction and relation schema definition at the sentence level, whereas AutoKG and KBTE operate at the document chunk level.\\

\noindent\textbf{Re-DocRED (\uppercase\expandafter{\romannumeral2})}\quad
We conducted experiments on the "Microsoft Windows" subset within the Re-DocRED dataset. This dataset presents greater challenges by re-annotating false negative entities in the original DocRED dataset. As demonstrated by Zhu et al., LLMs fail to surpass fully supervised small models on such datasets, \textbf{usually achieving F1 scores around 20} \cite{AutoKG}.
The observed discrepancy can be attributed to the fundamental distinction that fully supervised approaches leverage human-annotated ground truth labels during training. In contrast, LLMs do not learn potential relation sets from training data in advance, thereby resulting in a broader spectrum of relation types. Despite this inherent difference, \name demonstrates significantly superior performance in terms of precision and recall metrics compared to all baseline methods, notably achieving an F1 score approaching 30.

\subsubsection{Private Corpora}
We further conducted evaluations on the IMS documentation dataset, with the results presented in Table \ref{private_res}. As previously established, there are little publicly available reference about this system on the Internet. Therefore, constructing a high-quality knowledge graph from this dataset necessitates full utilization of contextual knowledge within the documents.

Generally, consistent with the public dataset, \name achieved superior precision and recall number. Compared to Prometheus documentation, the IMS documentation exhibits less extensive content and relatively simpler concepts, from which \name identified 41 entity types.

In terms of accuracy, AutoKG exhibits significant performance degradation compared to public datasets, while the other three methods demonstrate improvements. This occurs because AutoKG retrieves numerous irrelevant concepts (e.g., hydrogen energy companies) when attempting to acquire IMS-related knowledge by online retrieval. These concepts unrelated to monitoring systems lead to model hallucinations, substantially reducing the accuracy of tuple extraction. Both EDC and KBTE incorporate mechanisms for context-aware knowledge assistance to facilitate factual comprehension, thereby achieving higher accuracy on these relatively simple corpora. \name maintains superior accuracy by comprehensively evaluating the optimal order of document access, enabling more efficient utilization of knowledge contexts.

As for recall number, as previously noted, this dataset contains fewer knowledge concepts, resulting in generally lower average tuple recall numbers compared to the Prometheus dataset. Nevertheless, \name extracted 4,984 valid entities, ensuring sufficient recalled triples.

\begin{table}[t]
\renewcommand{\arraystretch}{1.2} 
\captionsetup{skip=5pt}
\setlength{\tabcolsep}{5pt} 
\caption{Evaluation results on IMS Documentation dataset} \label{private_res}
\begin{tabular}{|c|c|c|c|c|c|c|}
\hline
\rowcolor[HTML]{656565} 
\cellcolor[HTML]{656565}{\color[HTML]{FFFFFF} Model} &
\multicolumn{3}{c|}{\cellcolor[HTML]{656565}{\color[HTML]{FFFFFF} Meta-Llama-3.1-70B-Instruct}} &
\multicolumn{3}{c|}{\cellcolor[HTML]{656565}{\color[HTML]{FFFFFF} DeepSeek-R1-Distill-Qwen-14B}}
\\ 
\hline

\rowcolor[HTML]{EFEFEF} 
%\cellcolor[HTML]{EFEFEF}{\color[HTML]{000000} \diagbox{Method}{Metric}} &
Method\textbf{\textbackslash}Metric &
\cellcolor[HTML]{EFEFEF}{\color[HTML]{000000} Precision} &
\cellcolor[HTML]{EFEFEF}{\color[HTML]{000000} \begin{tabular}[c]{@{}c@{}}Recall\\ Number\end{tabular}} &
\cellcolor[HTML]{EFEFEF}{\color[HTML]{000000} \begin{tabular}[c]{@{}c@{}}Average\\ Recall\\ Number\end{tabular}} &
\cellcolor[HTML]{EFEFEF}{\color[HTML]{000000} Precision} &
\cellcolor[HTML]{EFEFEF}{\color[HTML]{000000} \begin{tabular}[c]{@{}c@{}}Recall\\ Number\end{tabular}} &
\cellcolor[HTML]{EFEFEF}{\color[HTML]{000000} \begin{tabular}[c]{@{}c@{}}Average\\ Recall\\ Number\end{tabular}} \\ \hline

EDC &
73.2\% &
1852 &
40.3 &
77.0\% &
2293 &
49.8 \\ \hline 

AutoKG &
56.7\% &
2296 &
49.9 &
64.3\% &
1424 &
31.0 \\ \hline 

KBTE &
80.4\% &
1637 &
35.6 &
74.0\% &
1067 &
23.2 \\ \hline 

LKD-KGC(Ours) &
\textbf{89.6\%} &
\textbf{2672} &
\textbf{58.1} &
\textbf{88.9\%} &
\textbf{2764} &
\textbf{60.1} \\ \hline 
\end{tabular}
\end{table}

\newpage

\section{Conclusion}
This paper proposes LKD-KGC, a novel framework for unsupervised domain-specific KG construction. By leveraging LLM-driven knowledge dependency parsing and autoregressive schema generation, LKD-KGC autonomously infers hierarchical corpora processing order and integrates cross-document contexts to build entity schemas. Experimental results on public and private datasets demonstrate its superiority over state-of-the-art methods in both precision and recall, validating its capability to handle domain-specific complexities. 
%% Note that this preceding line implies that you store your BibTeX references in a file called 'mybibliography.bib'. If you instead store your references in a file with a different name, for instance 'references.bib', the preceding line should read '\bibliography{references}'. Whatever you do, DO NOT put the file name extension .bib inside the \bibliography command; this will trip up LaTeX compilers. 
%
% If you do not want to use BibTeX, you can also type up the bibliography exactly as you see fit, using the following structure:

\bibliographystyle{splncs04}
\bibliography{main}
\end{document}